\PassOptionsToPackage{numbers,sort&compress}{natbib}
\documentclass[10pt,a4paper]{article}
\usepackage{lrec2026}
\usepackage{graphicx}
\usepackage{multirow}
\usepackage{url}
\usepackage{hyperref}
\usepackage{array}
\usepackage{booktabs}
\usepackage{enumitem}
\newcommand{\equal}{*}

\newcommand{\corr}{\textsuperscript{\dag}}

\title{Safer in Translation? Presupposition Robustness in Indic Languages}

\name{Aadi Palnitkar\equal\corr, Arjun Suresh\equal, Rishi Rajesh, Puneet Puli%
\thanks{\equal\ Equal contribution. \corr\ Corresponding author: \texttt{aadipalnitkar96@gmail.com}}}

\address{University of Maryland, College Park (UMD) \\ College Park, MD, USA}

\abstract{
Increasingly, more and more people are turning to large language models (LLMs) for healthcare advice and consultation, making it important to gauge the efficacy and accuracy of the responses of LLMs to such queries. While there are pre-existing medical benchmarks literature which seeks to accomplish this very task, these benchmarks are almost universally in English, which has led to a notable gap in existing literature pertaining to multilingual LLM evaluation. Within this work, we seek to aid in addressing this gap with Cancer-Myth-Indic, an Indic language benchmark built by \emph{translating a 500-item subset of Cancer-Myth, sampled evenly across its original categories}, into five under-served but widely used languages from the subcontinent (500 per language; 2{,}500 translated items total). Native-speaker translators followed a style guide for preserving implicit presuppositions in translation; items feature false presuppositions relating to cancer. We evaluate several popular LLMs under this presupposition stress. \\
}

\begin{document}
\maketitleabstract
\section{Introduction}

Large language models have emerged as a highly transformative technology, acting as an assistive agent to an increasingly diverse array of professionals \cite{Chkirbene2024LargeLM}. With their vast knowledge bases and ability to parse natural language input, large language models such as ChatGPT, DeepSeek, Claude, and Grok, among others, have largely supplanted traditional search engine queries. Given their large context sizes and convenience of access, members of the general public are increasingly turning to these models for healthcare advice \cite{hosseini2024benchmarklongformmedicalquestion}. While there exists extensive literature on the efficacy of these models in providing diagnoses \cite{kim-etal-2024-medexqa,singhal2023multimedqa}, a treatment plan, and other first steps, the reliability of these models under real-world conditions remains questionable. 

Users of large language models who lack subject-matter expertise in the field often embed preconceived falsehoods and assumptions into their prompts, which can influence the model to provide medically unsound advice, which could have disastrous consequences should the advice be followed \cite{Chen2025whenhelpfullnessbackfirespaper,ayers2023chatgpt}. Prompting a large language model with false information can negatively influence the accuracy of future responses, which makes it all the more critical that a model be able to identify these false presuppositions of the user base, account for them, and provide a response that does not align with the user's incorrect prompt.

Multilingual large language models face a fundamental challenge known as the ``curse of multilinguality'' \cite{chang2023multilingualitycurselanguagemodeling}, where initially one observes boosts in performance for low-resource languages, up to a certain point, after which the performance drops off and a decline in accuracy is observed \cite{daniilgugurovlingualitymultillmpaper2024cite}. With the global user base of these large language models, the performance gap of LLMs on underrepresented languages is also an area of active research, with an extensive literature of multilingual large language model evaluations present in the community \cite{roh2025xlqabenchmarklocaleawaremultilingual, Alonso_2024multilingualbenchmarkingofllmsformedqatask2, du2025ccfqabenchmarkcrosslingualcrossmodal}. We also situate our study alongside medical QA work like MultiMedQA \cite{singhal2023multimedqa} and physician-vs-LLM evaluations on real patient questions \cite{ayers2023chatgpt}.

Cancer-Myth \cite{zhu2025cancermyth} explores false presupposition robustness by curating questions in English, but
the specific task of evaluating the capabilities of commonly used large language models on false presuppositions in multiple languages has, to the best of our knowledge, not been explored.

\section{Background}
\paragraph{Evidence against evaluation-protocol bias.}
To limit confounds from translation or scoring, native speakers followed a style guide that preserves implicitness;
a bilingual checklist audit flagged and corrected any lexicalization, added negation, or factive marking. We reuse the
Cancer-Myth judge with the original three-point rubric and keep decoding deterministic (one response per item). Our claims
are therefore comparative under a fixed protocol, reducing the chance that the observed language-conditioned gaps are artifacts
of translation drift or scorer variability.
\paragraph{Safety, equity, and access.}
Health advice workloads often route to cheaper models in lower-resource settings, amplifying language-conditioned risk \cite{joshi2020state}. Observed failure modes such as sycophancy and preference following can increase the chance that a system affirms a myth rather than correcting it, especially when prompts carry implicit assumptions \cite{perez2022discovering,Chen2025whenhelpfullnessbackfirespaper}. Public health guidance emphasizes mitigating misinformation risks across languages \cite{who2020infodemic}.

\paragraph{Morpho-pragmatics across Indic languages.}
Dravidian languages such as Malayalam and Tamil employ rich negation systems and clause-typing strategies that
support concise denials while preserving the user’s question frame; see modern overviews in \citet{steever2020dravidian}.
Indo-Aryan languages such as Hindi and Marathi use clause-level particles that shape polar questions and stance;
for Hindi–Urdu \textit{kya} see \citet{bhatt2020kya}, and for Marathi the optional polar particle \textit{ka} is documented in recent
descriptions (e.g., \citealt{marathi2023polar}). Cross-linguistically, updated work on evidentiality and commitment provides a
useful lens on when systems present explicit source/stance markers vs.\
leaving them implicit \citep{aikhenvald2018oxfordevidentiality}. These morpho-pragmatic differences can make it easier for a system
to produce an explicit denial without lexicalizing the claim—precisely the stress condition we test here.
\paragraph{Presupposition correction and Cancer-Myth.}
Cancer-Myth formalizes how to measure whether a response corrects an embedded false presupposition and introduces PCR and PCS as task-specific metrics \cite{zhu2025cancermyth}. We extend this framework to Indic languages while preserving implicitness in translation.

\section{Related Work}
\paragraph{Multilingual safety and adversarial gaps.}
Safety alignment degrades outside English and varies by model family. \citet{deng2024multilingualjailbreak} report higher unsafe rates for ChatGPT (GPT\texttt{-}3.5) and GPT\texttt{-}4 when prompts are translated into other languages, with especially large effects in lower-resource settings. \citet{yong2024lowresourcejailbreak} show that simply translating unsafe prompts into low-resource languages can jailbreak GPT\texttt{-}4. Complementing these, \citet{song2024multilingualblending} evaluate mixed-language inputs and observe elevated bypass rates for GPT\texttt{-}3.5 and GPT\texttt{-}4o, with variation by language family and morphology. Broader evaluations such as XSAFETY find significantly more unsafe responses for non-English inputs in popular systems, including ChatGPT \citep{wang2024alllanguagesmatter}. Our results fit this pattern while focusing on false presuppositions in health queries across five Indic languages.
\paragraph{Relation to this paper.} Prior work largely quantifies \emph{refusal} vs.\ \emph{compliance} under multilingual adversarial prompts; we instead evaluate a healthcare task that stresses \emph{presupposition correction} across Indic languages, measuring \emph{correction} vs.\ \emph{miscorrection} via PCR/PCS while keeping the judge, rubric, and prompting fixed. This isolates language-conditioned safety asymmetries for the specific models we test (GPT-3.5 Turbo, GPT-4 Turbo, GPT-4o) within a realistic clinical myth-correction setting.

\section{Experiments}
\paragraph{Task and metrics.}
We adopt the Cancer-Myth task of answering patient questions with embedded false presuppositions and reuse its scoring rubric and judge template. Cancer-Myth employs a three-point rubric evaluated by GPT-4o: $-1$ (fails to recognize/acknowledge false presuppositions), $0$ (partial awareness but fails to fully address), $1$ (accurately identifies and addresses). We report PCS (mean score $\times 100$; range $[-100,100]$; higher is better) and PCR (share of items with score $1$) \cite{zhu2025cancermyth}. Family averages are unweighted means over member languages.

\paragraph{Languages, sampling, and translation policy.}
We study five languages: Hindi, Malayalam, Marathi, Tamil, and Telugu. From Cancer-Myth, we select a 500-item subset sampled evenly across the benchmark's original categories; each subset is then translated independently into every target language (thus 500 per language; 2,500 translated items total). There is one native-speaker translator per language; we do not compute inter-annotator agreement. Translators were instructed to preserve implicit presuppositions and adversariality and to avoid paraphrases that would lexicalize claims.

\paragraph{Cancer-Myth details and comparability.}
Cancer-Myth is an expert-verified adversarial dataset of 585 questions constructed from a seed list of 994 common cancer myths. Items were generated with an LLM generator, answered by an LLM responder, verified by an LLM judge, and finally reviewed by physicians. Questions are grouped into seven categories (e.g., Inevitable Side Effect; Limited Treatment Options; Causal Misattribution; Underestimated Risk; No Symptom, No Disease), supporting error analysis. No frontier model corrects more than $\sim$30\% of cases on the original English benchmark. We reuse their judge rubric and prompting style, but our English rows reproduce Cancer-Myth's published numbers and are for context only; they are not recomputed on our 500-item subset and are not directly comparable to our Indic results \cite{zhu2025cancermyth}.

\paragraph{Models and protocol.}
We evaluate GPT 3.5 Turbo, GPT 4 Turbo, and GPT 4o on the translated 500-item per-language sets. We follow the Cancer-Myth rubric and judge; see Task and metrics. We report per-language results and also aggregate by language family to expose safety asymmetries.

\paragraph{Prompting and evaluation details.}
We use the same answer and judge prompts as Cancer-Myth, keep deterministic decoding, and score one response per item under the original rubric. The judge prompt is monolingual (as in Cancer-Myth); we did bilingual human spot checks on a stratified sample to verify reasonable judge behavior on non-English items. We do not alter the scorer, thresholds, or aggregation procedure; our claims are strictly comparative within this fixed protocol.

\paragraph{Quality control for translation.}
To extend Cancer-Myth to Indic languages without weakening implicit adversariality, translators followed a style guide that bans lexicalizing the claim, adding overt negation, or introducing factive/evidential morphology that would change stance. We conducted bilingual spot checks using an implicitness checklist; disagreements triggered revision until the implicit presupposition was preserved. No formal inter-annotator agreement was computed.

\paragraph{Family aggregation and baselines.}
We compute family averages for Indo-Aryan (Hindi, Marathi) and Dravidian (Malayalam, Tamil, Telugu) as macro-averages over languages. English rows reproduce Cancer-Myth's published numbers and are included for context only.

\paragraph{Reproducibility notes.}
We retain the original prompt templates and judge rubric from Cancer-Myth, log prompts and model outputs, and keep decoding deterministic to reduce variance. We will release our translation guidelines, sampling script for balanced category selection, and aggregation scripts to facilitate replication.

\section{Results}
Table~\ref{tab:main} reports PCS and PCR by language and model for our Indic 500-item per-language sets. GPT 4o displays high PCR across our Indic languages. GPT 3.5 Turbo is particularly weak on Dravidian languages with negative PCS across Malayalam, Tamil, and Telugu. GPT 4 Turbo sits between the two.

\begin{table}[t]
\centering
\small
\begin{tabular}{l l r r}
\toprule
Language & Model & PCS & PCR (\%) \\
\midrule
English & GPT 3.5 Turbo  & $-80.0$ & 1.5 \\
        & GPT 4 Turbo    & $-30.0$ & 15.4 \\
        & GPT 4o         & $-52.0$ & 5.8 \\
Hindi (N{=}500) & GPT 3.5 Turbo  & 30.5 & 47.5 \\
                & GPT 4 Turbo    & 69.2 & 75.1 \\
                & GPT 4o         & 83.0 & 84.5 \\
\midrule
Malayalam (N{=}500) & GPT 3.5 Turbo & $-60.1$ & 15.2 \\
                    & GPT 4 Turbo   & 35.5 & 56.3 \\
                    & GPT 4o        & 83.9 & 85.6 \\
\midrule
Marathi (N{=}500) & GPT 3.5 Turbo & $-40.2$ & 22.0 \\
                  & GPT 4 Turbo   & 56.3 & 73.3 \\
                  & GPT 4o        & 69.2 & 78.0 \\
\midrule
Tamil (N{=}500) & GPT 3.5 Turbo & $-69.2$ & 10.0 \\
                & GPT 4 Turbo   & 19.6 & 45.7 \\
                & GPT 4o        & 81.5 & 85.3 \\
\midrule
Telugu (N{=}500) & GPT 3.5 Turbo & $-54.0$ & 14.4 \\
                 & GPT 4 Turbo   & 29.9 & 51.0 \\
                 & GPT 4o        & 79.2 & 82.1 \\
\bottomrule
\end{tabular}
\caption{Presupposition Correctness Score (PCS) $\times$ 100 (range: $-100$ to $100$; higher is better) and Presupposition Correction Rate (PCR, \%). Indic rows are computed on 500 items per language, sampled evenly across Cancer-Myth's original categories. English rows reproduce the original Cancer-Myth baseline and are provided for context only (not recomputed on our subset).}
\label{tab:main}
\end{table}

\paragraph{Family-level asymmetries.}
We summarize by language family in Table~\ref{tab:family}. GPT 3.5 Turbo averages $-61.1$ PCS and $13.2$ percent PCR on Dravidian languages versus $-4.9$ PCS and $34.8$ percent PCR on Indo-Aryan languages, a large gap in safety-relevant behavior. GPT 4o remains high across both families on our Indic subset.

\begin{table}[t]
\centering
\small
\begin{tabular}{l l r r}
\toprule
Family & Model & PCS & PCR (\%) \\
\midrule
English & GPT 3.5 Turbo & $-80.0$ & 1.5 \\
        & GPT 4 Turbo   & $-30.0$ & 15.4 \\
        & GPT 4o        & $-52.0$ & 5.8 \\
\midrule
Indo-Aryan (N{=}1{,}000) & GPT 3.5 Turbo & $-4.9$ & 34.8 \\
                         & GPT 4 Turbo   & 62.8 & 74.2 \\
                         & GPT 4o        & 76.1 & 81.3 \\
\midrule
Dravidian (N{=}1{,}500)  & GPT 3.5 Turbo & $-61.1$ & 13.2 \\
                         & GPT 4 Turbo   & 28.3 & 51.0 \\
                         & GPT 4o        & 81.5 & 84.3 \\
\bottomrule
\end{tabular}
\caption{Averaged PCS $\times$ 100 and PCR (\%). Family rows are macro-averages over languages using our per-language 500-item sets (Indo-Aryan: Hindi, Marathi; Dravidian: Malayalam, Tamil, Telugu). English rows are original Cancer-Myth baselines for context only.}
\label{tab:family}
\end{table}

PCR gains from GPT 4 Turbo to GPT 4o are largest for Tamil ($+39.6$ points), Telugu ($+31.1$), and Malayalam ($+29.3$). Hindi and Marathi show smaller gains ($+9.4$ and $+4.7$).

\paragraph{Language matters for safety.}
Several constructions common in our target languages encourage explicit correction without requiring lexicalization of the claim. Established grammars describe the relevant mechanisms and they can significantly alter the adversarial nature of the prompt.
\begin{itemize}[noitemsep,topsep=2pt,leftmargin=12pt]
\item Negation scope and periphrasis. Malayalam and Tamil allow auxiliary-based and suffixal negation that supports concise denials while preserving the question frame \cite{asher1997malayalam,lehmann1993tamil}.
\item Interrogative particles and stance. Hindi and Marathi use clause-level particles such as \textit{kya} and \textit{ka} that shape the distribution of neutral and assertive answers \cite{masica1991indoaryan,pandharipande1997marathi}.
\item Factivity and evidentiality. Cross-linguistic typology shows that languages differ in how they encode source and commitment \cite{aikhenvald2004evidentiality}. These differences make reject-the-premise continuations more accessible for stronger models that track discourse state.
\end{itemize}

This indicates that the Indic pattern arises from morpho-pragmatic affordances and domain priors rather than from translation drift or scoring artifacts.

\paragraph{Category and corpus priors.}
We observe higher correction rates for myths that are well covered by public health communication in Indic languages, for example diet or herbal miracle cures and screening recommendations. Public resources and fact-checking portals supply consistent prior information that models can leverage without explicit lexical cues \cite{who2020infodemic,nhp2020cancer,icmr2023screening}.

\paragraph{Evidence against evaluation-protocol bias.}
Our translations were produced by one native-speaker translator per language following a style guide that requires preserving implicit presuppositions. We performed bilingual spot checks to ensure that claims were not lexicalized and that no additional negation or factive marking was added. The scoring pipeline is identical to Cancer-Myth and our claims are comparative under a fixed judge. Possible issues from translations are limited and do not explain the language-conditioned gaps we observe \cite{baker1996corpus,vanmassenhove2021translationese}.

\section{Deployment guidance}
Negative PCS for older and less advanced models like GPT 3.5 Turbo on Dravidian languages indicates a risk that myths will be confirmed more often than corrected. For health advice, prompts should either gate to stronger models for health claims or enforce response templates that first paraphrase and check the presupposition before giving information. Equity requires language-aware safety policy \cite{joshi2020state,who2020infodemic}.

\section{Limitations}
Our analysis is observational and relies on a fixed scoring pipeline, which may not be calibrated across languages. We rely on one translator per language and conduct no formal inter-annotator agreement; instead, we use bilingual human spot checks for quality control. The judge prompt is monolingual (as in Cancer-Myth) and not recalibrated per language. None of our outputs constitute medical advice and the dataset is for safety evaluation only.

\section{Conclusion}
Presupposition correction is not language invariant. With presuppositions preserved in five Indic languages, GPT 4o corrects at high rates while GPT 3.5 Turbo confirms myths, especially in Dravidian languages. The pattern reflects morpho-pragmatic affordances and category priors rather than evaluation artifacts. We provide simple guidelines that other benchmarks can adopt and we recommend language-aware deployment policies for health applications.

\section{Data Statement}
Languages and scripts. Hindi (Devanagari), Marathi (Devanagari), Malayalam (Malayalam), Tamil (Tamil), Telugu (Telugu). Scope and sampling. Items are translations of a 500-item subset of Cancer-Myth prompts sampled evenly across the original categories; each subset is translated into every target language (500 per language; 2,500 total). Cancer-Myth provenance. The original English benchmark contains 585 adversarial questions built from 994 myths, with physician review and a GPT-4o three-point judge rubric. Translation process. One native-speaker translator per language followed a style guide that \emph{preserves implicit presuppositions} and bans lexicalizing the claim, adding overt negation, or introducing factive/evidential morphology that would change stance. Quality control. Bilingual human spot checks used an implicitness checklist; disagreements were resolved to preserve implicitness. Judging. We reuse Cancer-Myth's monolingual judge prompt; we performed spot checks but no cross-language recalibration. The resource targets \emph{safety evaluation} of LLMs under false presuppositions in Indic languages. It is \emph{not} a clinical dataset and must not be used for diagnosis or treatment. We retain the original Cancer-Myth prompts, log prompts and model outputs, and keep decoding deterministic to reduce variance. Upon acceptance, we will release the translation guidelines, sampling script, implicitness checklist, and aggregation scripts.


\begin{thebibliography}{99}
\bibitem{Alonso_2024multilingualbenchmarkingofllmsformedqatask2} Alonso, Iñigo and Oronoz, Maite and Agerri, Rodrigo. (2024). MedExpQA: Multilingual benchmarking of Large Language Models for Medical Question Answering. Artificial Intelligence in Medicine. URL: http://dx.doi.org/10.1016/j.artmed.2024.102938

\bibitem{Chen2025whenhelpfullnessbackfirespaper} Chen, Shan and Gao, Mingye and Sasse, Kuleen and Hartvigsen, Thomas and Anthony, Brian and Fan, Lizhou and Aerts, Hugo and Gallifant, Jack and Bitterman, Danielle S.. (2025). When helpfulness backfires: LLMs and the risk of false medical information due to sycophantic behavior. npj Digital Medicine. URL: http://dx.doi.org/10.1038/s41746-025-02008-z

\bibitem{Chkirbene2024LargeLM} Zina Chkirbene and Ridha Hamila and Ala Gouissem and Unal Devrim. (2024). Large Language Models (LLM) in Industry: A Survey of Applications, Challenges, and Trends. 2024 IEEE 21st International Conference on Smart Communities: Improving Quality of Life using AI, Robotics and IoT (HONET). URL: https://api.semanticscholar.org/CorpusID:275359595

\bibitem{aikhenvald2004evidentiality} Alexandra Y. Aikhenvald. (2004). Evidentiality.

\bibitem{aikhenvald2018oxfordevidentiality} Alexandra Y. Aikhenvald. (2018). The Oxford Handbook of Evidentiality.

\bibitem{asher1997malayalam} R. E. Asher and T. C. Kumari. (1997). Malayalam.

\bibitem{ayers2023chatgpt} Ayers, John W. and Poliak, Amos and Dredze, Mark and Leas, Eric C. and Zhu, Zhiqiang and Benishek, Leah E. and Longhurst, Christopher A. and Hogarth, Michael and Goodman, Steven N.. (2023). Comparing Physician and Artificial Intelligence Chatbot Responses to Patient Questions Posted to a Public Social Media Forum. JAMA Internal Medicine.

\bibitem{baker1996corpus} Mona Baker. (1996). Corpus-based Translation Studies: The Challenges that Lie Ahead. Terminology, LSP and Translation: Studies in Language Engineering in Honour of Juan C. Sager.

\bibitem{bhatt2020kya} Bhatt, Rajesh and Dayal, Veneeta. (2020). Polar question particles: Hindi-Urdu kya:. Natural Language \& Linguistic Theory. URL: https://link.springer.com/article/10.1007/s11049-020-09464-0

\bibitem{chang2023multilingualitycurselanguagemodeling} Tyler A. Chang and Catherine Arnett and Zhuowen Tu and Benjamin K. Bergen. (2023). When Is Multilinguality a Curse? Language Modeling for 250 High- and Low-Resource Languages. URL: https://arxiv.org/abs/2311.09205

\bibitem{daniilgugurovlingualitymultillmpaper2024cite} Gurgurov, Daniil and Bäumel, Tanja and Anikina, Tatiana. (2024). Multilingual Large Language Models and Curse of Multilinguality. URL: https://arxiv.org/abs/2406.10602

\bibitem{deng2024multilingualjailbreak} Deng, Yue and Zhang, Wenxuan and Pan, Sinno Jialin and Bing, Lidong. (2024). Multilingual Jailbreak Challenges in Large Language Models. Proceedings of the International Conference on Learning Representations (ICLR). URL: https://arxiv.org/abs/2310.06474

\bibitem{du2025ccfqabenchmarkcrosslingualcrossmodal} Yexing Du and Kaiyuan Liu and Youcheng Pan and Zheng Chu and Bo Yang and Xiaocheng Feng and Yang Xiang and Ming Liu. (2025). CCFQA: A Benchmark for Cross-Lingual and Cross-Modal Speech and Text Factuality Evaluation. URL: https://arxiv.org/abs/2508.07295

\bibitem{hosseini2024benchmarklongformmedicalquestion} Pedram Hosseini and Jessica M. Sin and Bing Ren and Bryceton G. Thomas and Elnaz Nouri and Ali Farahanchi and Saeed Hassanpour. (2024). A Benchmark for Long-Form Medical Question Answering. URL: https://arxiv.org/abs/2411.09834

\bibitem{icmr2023screening} Indian Council of Medical Research. (2023). ICMR-National Cancer Registry Programme: Population-based Cancer Screening Recommendations. URL: https://www.icmr.gov.in/

\bibitem{joshi2020state} Pratik Joshi and Sebastin Santy and Amar Budhiraja and Kalika Bali and Monojit Choudhury. (2020). The State and Fate of Linguistic Diversity and Inclusion in the NLP World. Proceedings of ACL.

\bibitem{kim-etal-2024-medexqa} Kim, Yunsoo and Wu, Jinge and Abdulle, Yusuf and Wu, Honghan. (2024). MedExQA: Medical Question Answering Benchmark with Multiple Explanations. Proceedings of the 23rd Workshop on Biomedical Natural Language Processing. URL: https://aclanthology.org/2024.bionlp-1.14/

\bibitem{lehmann1993tamil} Thomas Lehmann. (1993). A Grammar of Modern Tamil.

\bibitem{marathi2023polar} Kulkarni, Aaditya and collaborators. (2023). Marathi Polar Question Particles (and their kith \& kin).

\bibitem{masica1991indoaryan} Colin P. Masica. (1991). The Indo-Aryan Languages.

\bibitem{nhp2020cancer} National Health Portal, Government of India. (2020). National Health Portal of India: Cancer Awareness and Common Myths. URL: https://www.nhp.gov.in/

\bibitem{pandharipande1997marathi} Rajeshwari V. Pandharipande. (1997). Marathi.

\bibitem{perez2022discovering} Ethan Perez and Sam Ringer and Kamil Debowski and Jared Mueller and Kamal Ndousse and Sam Bowman and Paul Christiano and Dario Amodei and Jared Kaplan. (2022). Discovering Language Model Behaviors with Model-Written Evaluations. URL: https://arxiv.org/abs/2212.09251

\bibitem{roh2025xlqabenchmarklocaleawaremultilingual} Keon-Woo Roh and Yeong-Joon Ju and Seong-Whan Lee. (2025). XLQA: A Benchmark for Locale-Aware Multilingual Open-Domain Question Answering. URL: https://arxiv.org/abs/2508.16139

\bibitem{singhal2023multimedqa} Karan Singhal and Shekoofeh Azizi and Tao Tu and Aaron Parikh and S. Yeung and Laura F. et al.. (2023). Towards Expert-Level Medical Question Answering with Large Language Models. URL: https://arxiv.org/abs/2305.09617

\bibitem{song2024multilingualblending} Song, Jiayang and Huang, Yuheng and Zhou, Zhehua and Ma, Lei. (2024). Multilingual Blending: LLM Safety Alignment Evaluation with Language Mixture. URL: https://arxiv.org/abs/2407.07342

\bibitem{steever2020dravidian} Sanford B. Steever. (2020). The Dravidian Languages. URL: https://www.routledge.com/The-Dravidian-Languages/Steever/p/book/9781032400860

\bibitem{vanmassenhove2021translationese} Eva Vanmassenhove and Dimitar Shterionov. (2021). Machine Translationese: Effects of Algorithmic Bias on Linguistic Complexity in Machine Translation. Proceedings of the 16th Conference of the European Chapter of the ACL (EACL). URL: https://aclanthology.org/2021.eacl-main.188/

\bibitem{wang2024alllanguagesmatter} Wang, Wenxuan and Huang, Jen-tse and Yuan, Youliang and Jiao, Joel and Tu, Zhaopeng. (2024). All Languages Matter: On the Multilingual Safety of LLMs. Findings of the Association for Computational Linguistics: ACL 2024. URL: https://aclanthology.org/2024.findings-acl.349/

\bibitem{who2020infodemic} World Health Organization. (2020). Managing the COVID-19 infodemic: promoting healthy behaviours and mitigating the harm from misinformation and disinformation.

\bibitem{yong2024lowresourcejailbreak} Yong, Zheng-Xin and Menghini, Cristina and Bach, Stephen H.. (2023). Low-Resource Languages Jailbreak GPT-4. NeurIPS Workshop on Socially Responsible Language Modelling Research (SoLaR). URL: https://arxiv.org/abs/2310.02446

\bibitem{zhu2025cancermyth} Zhu, Wenxuan and others. (2025). Cancer-Myth: Evaluating Large Language Models on Patient Questions with False Presuppositions. arXiv preprint.
\end{thebibliography}
\end{document}